\def\year{2022}\relax
\documentclass[letterpaper]{article} 
\usepackage{aaai22}  
\usepackage{times}  
\usepackage{helvet}  
\usepackage{courier}  
\usepackage[hyphens]{url}  
\usepackage{graphicx} 
\usepackage{natbib}  
\usepackage{caption} 
\DeclareCaptionStyle{ruled}{labelfont=normalfont,labelsep=colon,strut=off} 
\usepackage{algorithm}
\usepackage{algorithmic}
\usepackage{hyperref}

\usepackage{newfloat}
\usepackage{listings}

\usepackage{times}
\usepackage{epsfig}
\usepackage{graphicx}
\usepackage{amsmath}
\usepackage{amssymb}
\usepackage{lipsum}

\usepackage{booktabs}
\usepackage{float}
\usepackage{hhline}
\usepackage{tabularx}

\usepackage{times}
\usepackage{epsfig}
\usepackage{graphicx}
\usepackage{amsmath}
\usepackage{amssymb}
\usepackage{esvect}

\usepackage{url}
\usepackage{microtype}
\usepackage{graphicx}
\usepackage{subfigure}
\usepackage{booktabs} 
\usepackage{xcolor}
\usepackage{amsthm}
\usepackage{multirow}
\usepackage{diagbox}
\usepackage{arydshln}
\usepackage{mathtools}
\usepackage{soul}
\usepackage{algorithm}
\usepackage{algorithmic}

\floatstyle{ruled}
\newfloat{listing}{tb}{lst}{}
\floatname{listing}{Listing}

\nocopyright

\title{Noise Estimation for Generative Diffusion Models}

\title{Noise Estimation for Generative Diffusion Models}
\author {
    Robin San Roman*\textsuperscript{\rm 1},
    Eliya Nachmani* \textsuperscript{\rm 2,3},
    Lior Wolf \textsuperscript{\rm 2}
}
\affiliations {
    \textsuperscript{\rm 1} École Normale Supérieure Paris-Saclay \\
    \textsuperscript{\rm 2} Tel-Aviv University\\
    \textsuperscript{\rm 3} Facebook AI Research\\
    sanroman.robin@gmail.com, enk100@gmail.com, wolf@cs.tau.ac.il
}

\usepackage{bibentry}

\begin{document}

\maketitle
{\let\thefootnote\relax\footnote{{*Equal contribution}}}
\begin{abstract}
Generative diffusion models have emerged as leading models in speech and image generation. However, in order to perform well with a small number of denoising steps, a costly tuning of the set of noise parameters is needed. In this work, we present a simple and versatile learning scheme that can step-by-step adjust those noise parameters, for any given number of steps, while the previous work needs to retune for each number separately. Furthermore, without modifying the weights of the diffusion model, we are able to significantly improve the synthesis results, for a small number of steps. Our approach comes at a negligible computation cost. 
\end{abstract}

\section{Introduction}

Deep generative models have seen a tremendous advancement in the past few years. The main successful architectures can be divided into two categories: (i) autoregessive models, such as VQ-VAE for images \cite{razavi2019generating} and Wavenet for speech \cite{oord2016wavenet}. (ii) non-autoregessive models, for example, StyleGAN \cite{karras2020analyzing} for vision application and WaveGAN \cite{donahue2018adversarial} for audio synthesis. 

An emerging class of non-autoregessive  models is the one of Denoising Diffusion Probabilistic Models (DDPM). Such methods use diffusion
models and denoising score matching in order to generate images \cite{ho2020denoising} and speech \cite{chen_wavegrad_2020}. The DDPM model learns to perform a diffusion process on a Markov chain of latent variables. The diffusion process transforms a data sample into Gaussian noise. During inference the reverse process is used, which is called the denoisng process. The inference procedure starts from Gaussian noise and iteratively refines the signal. This process is often conditioned on the class and attributes that one wishes to generate. 

In order to get high quality synthesis, a large number of denosing steps are used (i.e. $1000$ steps). To allow the process to converge to a high quality output with only a small number of denosing steps, a costly grid search is required in order to find a noise schedule that would produce high-fidelity results. 

In this paper, we propose a novel method for obtaining the noise schedule based on the conditioning data. The noise schedule that the proposed method produces leads to a high fidelity synthesis of samples, even for a small number of denosing steps.  Moreover, for a given amount of denosing steps, the proposed method obtains better results than the previous models, when their noise schedule is determined by a costly per-sample grid search for the optimal parameters.

Our method introduces a novel neural network that is able to monitor and control the denoising process. Instead of fixing in advance the steps of the reverse process that will be skipped, this network is able, by estimating the amount of noise in the data, to schedule the subsequent steps of the denoising process.

Our results are demonstrated on two major domains: vision and audio. In the first domain, the proposed method is shown to provide a better FID score for generated images, when the number of steps is restricted. For speech data, we show that the proposed method improves various measures, such as Perceptual Evaluation of Speech Quality (PESQ) and short-time objective intelligibility (STOI).

\section{Related Work}
Diffusion Probabilistic Models were first introduced in the seminal work of Sohl-Dickstein et al.~\cite{sohl2015deep}, who presented the idea of using an iterative neural diffusion process for destroying the structure of a given distribution, while learning the reverse neural diffusion process for restoring the structure in the data. It was shown that the proposed neural diffusion process can learn the data distribution in domains, such as images and time series. The main issue with the proposed neural diffusion process is that during training it requires up to thousands of iterative steps in order to learn the target distribution.

In~\cite{song2019generative}, a new generative model based on the score matching method~\cite{hyvarinen2005estimation} and Langevin dynamics was introduced. The proposed model estimates and samples the logarithm of the data density, which is the Stein score function~\cite{liu2016kernelized}. The proposed method achieves state of the art results for modeling the CIFAR-10 dataset.

The two ideas of (i) neural Diffusion Probabilistic Models and (ii) generative models based on score matching, were combined by the DDPM method of Ho et al.~\cite{ho2020denoising}. DDPM presents a generative model based on the neural diffusion process and applies score matching for image generation. Subsequently, in \cite{chen_wavegrad_2020} a generative neural diffusion process based on score matching was applied to speech generation, obtaining state of the art results in comparison to well-established methods, such as Wavenet \cite{oord2016wavenet}, Wavernn \cite{kalchbrenner2018efficient} and GAN-TTS \cite{binkowski2019high}. A parallel contribution presented high fidelity speech generation results using a different neural diffusion process~\cite{kong2020diffwave}.

One major limitation of the generative neural diffusion process is that in order to generate a high quality sample, one should use a large number of diffusion steps, e.g., a thousand steps are often used. Denoising Diffusion Implicit Models (DDIMs)~\cite{song2020denoising} is an acceleration for the denoising process. It employs non-Markovian diffusion processes, which leads to a faster sample procedure than other diffusion models. 

A further development in the score based generative models is to consider this process as a solution to a stochastic differential equation~\cite{song2020score}. This method achieves state of the art performance for unconditional image generation on CIFAR-10. An alternative approach trains an energy-based generative model using a diffusion process that is applied to increasingly noisy versions of a dataset~\cite{gao2020learning}, also presenting results on CIFAR-10.

 The recent TimeGrad model~\cite{rasul2021autoregressive} is a diffusion process for probabilistic time series forecasting, which was shown empirically to outperform Transformers~\cite{vaswani2017attention} and LSTMs~\cite{hochreiter1997long} on some datasets.  In another concurrent work, a multinomial diffusion process is learned by adding categorical noise to the process~\cite{hoogeboom2021argmax}. Competitive results are presented for image segmentation and language modeling.

\section{Background}

Denoising Diffusion Probabilistic Model (DDPM) are neural network that learn the gradients of the data log density $\nabla_y \log p(y)$:

\begin{equation}
    s(y) = \nabla_y \log p(y) 
\end{equation}

Given those gradients, one can then use Langevin Dynamics to sample from the probability iteratively

\begin{equation}\label{eq:langevin}
    \tilde y_{i+1} = \tilde y_i + \frac{\eta}{2} s(\tilde y_i) + \sqrt{\eta}z_i
\end{equation}

Where $\eta > 0$ is the step size and $z_i\sim \mathcal{N}(0, I)$. 

The formalization of Denoising Diffusion Probabilistic Models (DDPM) by Ho et al \cite{ho2020denoising} employs a parameterized Markov chain trained using variational inference, in order to produce samples matching the data after finite time. The transitions of this chain are learned to reverse a diffusion process. This diffusion process is defined by a Markov chain that gradually adds noise in the data with a noise schedule $\beta_1, \dots \beta_N$ and is defined as: 
\begin{equation}
    q(y_{1:N}|y_0) = \prod_{n=1}^{N} q(y_n|y_{n-1})\,,
\end{equation}
where N is the length of the diffusion process, and $y_N,...,y_n,y_{n-1},...,y_0$ is a sequence of latent variables with the same size as the clean sample $y_0$.

At each iteration, the diffusion process adds Gaussian noise, according to the noise schedule: 
\begin{equation}
     q(y_n|y_{n-1}) := \mathcal{N}(y_{n}; \sqrt{1 - \beta_n}y_{n-1}, \beta_n\mathbf{I})\,, 
\end{equation}
where $\beta_n,$ is the noise schedule as defined above. 

The diffusion process can be simulated for any number of steps with the closed formula: 

\begin{equation}\label{eq:closedy_n}
    y_n = \sqrt{\bar \alpha} y_0 + \sqrt{1 - \bar\alpha} \varepsilon\,,
\end{equation}
where $\alpha_i = 1 - \beta_i$, $\bar \alpha_n = \prod_{i=1}^n \alpha_i$ and $\varepsilon = \mathcal{N}(0,\mathbf{I})$.

One can use this to implement the DDPM training algorithm (Alg.\ref{alg:DDPM_train}) which is defined in \cite{chen_wavegrad_2020}. The input to the training algorithm is the dataset $d$. The algorithm samples $s$, $\bar \alpha$ and $\epsilon$. The noisy latent variable $y_s$ is calculated and fed to the DDPM neural network $\varepsilon_\theta$. A gradient descent step is taken in order to estimate the $\varepsilon$ noise with the DDPM network. By the end of the algorithm, the DDPM network can estimate the noise added during the diffusion process.

When $\sqrt{\bar \alpha}$ is close to $1$, the diffusion process adds a small amount of noise and when $\sqrt{\bar \alpha}$ is close to $0$, there are large amounts of noise that are added to the generation process. 

As mentioned in \cite{chen_wavegrad_2020}, sampling the noise level $\sqrt{\bar \alpha}$ in the uniform distribution $\mathcal{U}([0,1])$ gives poor empirical results. This is due to the fact that the network $\varepsilon_\theta$ would rarely be trained to fine tune good examples ($\sqrt{\bar \alpha}$ close to $1$).

Instead, one can sample $\sqrt{\bar \alpha}$ such that the distribution of the training examples match the forward process, i.e., there are an equal amount of training samples that correspond to every step of the diffusion process. 
The first step is to sample a state $n$ ($n\sim \mathcal{U}\{1,\dots N\}$ line 3) in the forward process and then sample the noise level using:

\begin{equation}\label{eq:noise_distribution}
    \sqrt{\bar\alpha} \sim \mathcal{U}[l_n, l_{n-1}]
\end{equation}

Where: 
\begin{equation}
\label{eq:l}
    l_0 = 1,\hspace{1mm} l_n = \sqrt{\prod_{i=0}^n 1-\beta_i}
\end{equation}

\begin{algorithm}[tb]
   \caption{DDPM training procedure}
   \label{alg:DDPM_train}
\begin{algorithmic}[1]
   \REPEAT
   \STATE $ y_0 \sim d(y_0)$
   \STATE $n \sim \mathcal{U}(\{1, ..., N\})$
   \STATE $\sqrt{\bar \alpha} \sim \mathcal{U}( \left[ l_{n-1}, l_n\right] )$
   \STATE $\varepsilon \sim \mathcal{N}(0, I)$
   \STATE $y_n = \sqrt{\bar \alpha}y_0 + \sqrt{1 - | \bar \alpha|}\varepsilon$
   \STATE Take gradient descent step on: \newline $\| \varepsilon - \varepsilon_{\theta}(y_s, x, \sqrt{\bar \alpha} )\|_1 $
   \UNTIL{converged}
\end{algorithmic}
\end{algorithm}

In Algorithm \ref{alg:DDPM_train} (line 7) the DDPM is trained to learn the noise $\varepsilon$ directly, instead of learning the Markov chain gradients. In \cite{ho2020denoising} the authors show that the following reparametrization leads to better empirical results:
\begin{equation}\label{eq:reparam}
    \varepsilon = -\sqrt{1-\bar\alpha_n}\nabla_{y_n}\log q(y_n|y_0) 
\end{equation}

The trained model $\varepsilon_{\theta}$ can be used to perform inference using a variation of Langevin dynamics (Eq. \ref{eq:langevin}). The following update from \cite{song2020denoising} is used to reverse a step of the diffusion process:
\begin{equation}\label{eq:DDPM_update}
y_{n-1} = \dfrac{y_n - \frac{1 - \alpha_n}{\sqrt{1- \bar\alpha_n}}\varepsilon_\theta(y_n, x, \sqrt{\bar \alpha_n})}{\sqrt{\bar\alpha_n}} + \sigma_n \varepsilon\,,
\end{equation}

where $\varepsilon$ is white noise. Ho et al.~\cite{ho2020denoising} showed that adding white noise of variance $\sigma_n^2 = \beta_t$ is optimal, when the inference procedure is initialized with gaussian noise ($y_N\sim \mathcal{N}(O, I)$).

One can use this update rule in Eq.~\ref{eq:DDPM_update} to sample from the data distribution, by starting from a Gaussian noise and then step-by-step reversing the diffusion process. Algorithm~\ref{alg:DDPM_inf} is used to sample with the network $\varepsilon_\theta$.

Since our experiments on images are unconditional, the network no longer needs the input $x$. The update equation that we use is defined in~\cite{song2020denoising}:
\begin{equation}\label{eq:DDIM_update}
    y_{n-1} = \sqrt{\bar \alpha_{n-1}} \hat y_{0,n} + \sqrt{1 - \bar\alpha_{n-1} - \tilde \sigma_n^2}\varepsilon_\theta(y_n, \bar\alpha_n) + \tilde \sigma_n\varepsilon\,,
\end{equation}
where $\hat y_{0,n} = \dfrac{y_n - \sqrt{1-\bar \alpha_n}\varepsilon_\theta(y_n, \bar\alpha_n)}{\sqrt{\bar \alpha_n}}$ is the prediction of $y_0$, $\varepsilon \sim \mathcal{N}(0, I)$ is white noise, and $\tilde \sigma$ is a new parameter of the generative process.

One can apply the rule of Eq.\ref{eq:DDIM_update} with $\tilde \sigma=0$, in which case no random noise is added. This makes the process deterministic and leads to the best results  in most of the scenarios~\cite{song2020denoising}.

\begin{algorithm}[tb]
   \caption{DDPM sampling algorithm}
   \label{alg:DDPM_inf}
\begin{algorithmic}[1]
  \STATE $y_N \sim \mathcal{N}(0, I) $
   \FOR{n= N, ..., 1}
   \STATE $ z \sim \mathcal{N}(0, I)$
   \STATE $\hat \varepsilon = \varepsilon_\theta(y_n, x, \sqrt{\bar\alpha_n})$
   \STATE $y_{n-1} = \frac{y_n - \frac{1 - \alpha_n}{\sqrt{1 - \bar \alpha_n}}\hat \varepsilon }{\sqrt{\alpha_n}}$ 
   \IF{$n \neq 1$}
   \STATE $y_{n-1} = y_{n-1} + \sigma_n z$ 
   \ENDIF
   \ENDFOR
   \STATE \textbf{return} $y_0$
\end{algorithmic}
\end{algorithm}

\begin{algorithm}[tb]
   \caption{$P_\theta$ training procedure}
   \label{alg:example}
\begin{algorithmic}[1]
   \REPEAT
   \STATE $ y_0 \sim q(y_0)$
   \STATE $s \sim \mathcal{U}(\{1, ..., N\})$
   \STATE $\sqrt{\bar \alpha} \sim \mathcal{U}( \left[ l_{s-1}, l_s\right] )$
   \STATE $\varepsilon \sim \mathcal{N}(0, I)$
   \STATE $y_s = \sqrt{\bar \alpha}y_0 + \sqrt{1 - | \bar \alpha|}\varepsilon$
   \STATE $\hat{\alpha} = P_{\theta}(y_s)$
   \STATE Take gradient descent step on: \newline $\| \log(1-\bar\alpha) - \log(1-\hat \alpha)\|_2 $
   \UNTIL{converged}
\end{algorithmic}
\end{algorithm}

\begin{algorithm}[t]
   \caption{Model inference procedure}
   \label{alg:alpha_inf}
\begin{algorithmic}[1]
  \STATE $N$ Number of iterations
  \STATE $y_N \sim \mathcal{N}(0, I) $
  \STATE $\mathbf{\alpha}, \mathbf{\beta}= $ initialNoiseSchedule()
   \FOR{n= N, ..., 1}
   \STATE $ z \sim \mathcal{N}(0, I)$
   \STATE $\hat \varepsilon = \varepsilon_\theta(y_n, \sqrt{\bar\alpha_n})$ or $\varepsilon_\theta(y_n, t)$ where $ \bar \alpha_n\in [l_t, l_{t-1}]$
   \STATE $y_{n-1} = \frac{y_n - \frac{1 - \alpha_n}{\sqrt{1 - \bar \alpha_n}}\hat \varepsilon }{\sqrt{\alpha_n}}$ 
   \IF{$n \in U$}
   \STATE $\hat \alpha = P_\theta(y_{n-1})$
   \STATE $\mathbf{\alpha}, \mathbf{\beta}, \tau = $ updateNoiseSchedule($\hat \alpha, n$)
   \ENDIF
   \IF{$n \neq 1$}
   \STATE $y_{n-1} = y_{n-1} + \sigma_n z$
   \ENDIF
   \ENDFOR
   \STATE \textbf{return} $y_0$
\end{algorithmic}
\end{algorithm}

\begin{figure*}[t]
    \centering
        \includegraphics[width=0.9\linewidth]{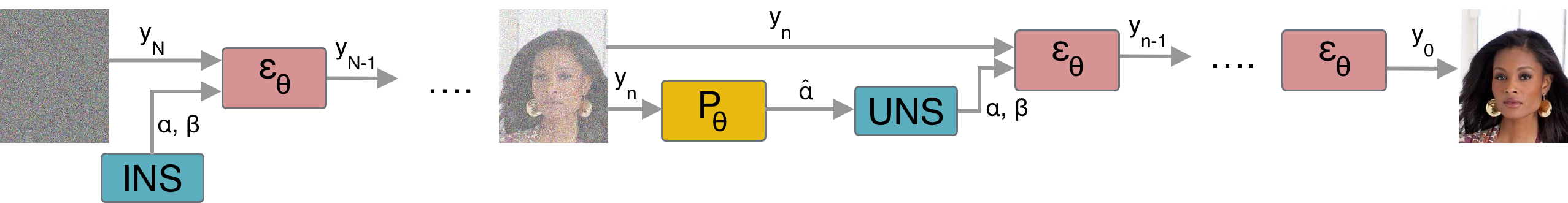}
    \caption{An overview of our generative process. INS and UNS are respectively the functions initializeNoiseSchedule() and updateNoiseSchedule($\hat\alpha$).}
    \label{fig:Our_process}
\end{figure*}

\section{Method}
We note that at training time the data is constructed in such a way (cf. Eq.~\ref{eq:closedy_n}) that we can feed the network $\varepsilon_\theta$ with the noisy data $y_n$ and with ground truth noise level $\sqrt{\bar\alpha}$. However, at a given inference step, the amount of noise in the data $y_n$ is unknown. In most methods, the conditioning used is a predefined one. As a result, the input conditioning given to the network $\varepsilon_\theta$ is not exploited at its full potential. 

This inexact analysis of the noise is especially detrimental when the number of steps in the generation process is small. In this case, the quality of the samples at intermediate states ($y_n$) varies widely. 

To solve this problem, we introduce a novel neural network $P_{\theta}$ that estimates the value of $\bar \alpha$, thus providing better conditioning for the diffusion network. The input to the neural network $P_\theta$ is the data $y_n$ generated at step $n$ and its output is the estimated noise level $\hat \alpha$. This network provides a continuous control signal to the generation process, by providing the DDPM network $\varepsilon_{\theta}$ with a conditioning signal that relates to the actual quality of the generation.

Figure \ref{fig:Our_process} depicts the generation process used. Similarly to  \cite{song2020denoising}, the idea is to use a pretrained DDPM ($\varepsilon_\theta$) from \cite{ho2020denoising} and skip some of the states in the process to shorten the generation time.  However, our model includes a Noise estimation model ($P_\theta$) to calculate, between denoising steps, adjustments of the noise schedule.


\subsection{Noise Level Estimation}
The network is trained with a similar procedure as Alg.~\ref{alg:DDPM_train}. The sampling of the noise level is done using the distribution described in Eq.~\ref{eq:noise_distribution}. Given the input $y_n$, the network $P_\theta$ estimates the noise level ($\bar\alpha$).

Empirically, we found that that the performance of the network in low noise situations is critical to the performance of our method. Indeed, the last steps of the generation process are responsible for the quality of the final sample. At those stages, the amount of noise is very small, and $\bar\alpha$ approaches $1$. (See experiments Fig. \ref{fig:p_theta_and_l1}(b)) 
We, therefore, design our regression loss on $ \bar \alpha$ : 
\begin{equation}\label{eq:loss}
    \mathcal{L}(\bar\alpha, \hat \alpha) = \| \log(1-\bar \alpha) - \log(1-\hat \alpha)\|_2
\end{equation}

This loss penalizes with higher cost the errors when close to 1 resulting in better network performances in this region.

\subsection{Noise Schedule Adjustments} \label{section:NSA}
    
We wish to use $\hat \alpha$, the output of the $P_\theta$, in order to adjust the noise schedule parameters. In the following, we present how obtain those parameters, assuming they follow either a linear or a Fibonacci distribution. This allows us to define the following function:

\begin{equation}
    \alpha, \beta = \text{updateNoiseSchedule}(\hat \alpha, n)   
\end{equation}

Where $\alpha, \beta$ are the set of noise parameters and $n$ is the number of remaining denoising steps. As we show, in order to define this function, it is sufficient to estimate the first parameters $\beta_0$ and the type of distribution. 

\subsubsection{Linear Schedule} \label{section:LNS}

In the general case, $\bar \alpha$ is given by:

\begin{equation} \label{eq:alpha_bar}
    \bar \alpha = \prod_{i=0}^{n-1} (1 - \beta_i)
\end{equation}

Since the values of the $\beta_i$ are typically between $10^{-6}$ and $10^{-2}$ we can use the Taylor approximation $log(1 - \beta_i) \approx - \beta_i$.
We can derive from Eq.~\ref{eq:alpha_bar}:

\begin{equation} \label{eq:approx}
    \log(\bar \alpha) = \sum_{i=0}^{n-1} \log(1 - \beta_i)\approx - \sum_{i=0}^{n-1} \beta_i
\end{equation}

Assuming the linear schedule, the expression of $\beta_i$ with respect to $i$ in the range $\{0, \dots n-1\}$ is:
\begin{equation}\label{eq:linear_betas}
     \beta_i = \beta_0 + ix\,,
\end{equation}
where $x$ is the step size. Therefore, we have: 

\begin{equation}\label{eq:log_app}
    \log(\bar \alpha) = - \left( \sum_{i=0}^{n-1} \beta_0 +ix \right)
\end{equation} 
and
\begin{equation}\label{eq:log_app2}
     x = -2 \frac{(\log(\bar \alpha) + n\beta_0)}{n(n-1)}
\end{equation}

Once $x$ is recovered, Eq.\ref{eq:linear_betas} provides us with the noise parameters required to perform the remaining denoising steps.

\subsubsection{Fibonacci Schedule} 
\label{FNS}
    
Chen et al.~\cite{chen_wavegrad_2020} employ a Fibonacci schedule for a low number of denoising steps:
\begin{equation}
    \beta_{i+2} = \beta_i + \beta_{i+1}
\end{equation}

We can find a closed form for this series given $\bar \alpha$ and $\beta_0$, which will allow us to compute all the terms. The homogenuous recurrent equation is:

\begin{equation}
    \beta_{i+2} - \beta_{i+1} - \beta_i = 0
\end{equation}

Thus, the series ($\beta_i$) is of the form: 
\begin{equation} 
    \beta_i = A  \varphi^i  + B \varphi'^i\,, 
\end{equation}   
where $\varphi$ and $\varphi'$ are the solutions of $x^2 - x - 1 = 0$

With a straightforward induction on $n$, one can show that:
\begin{equation}\label{eq:fib_sum}
    \forall n > 2, \beta_1 + \sum_{i=0}^{n-1}\beta_i = \beta_{n+1}
\end{equation}

Combining Eq.~\ref{eq:fib_sum} with the approximation of Eq.~\ref{eq:approx}, we obtain that $A$ is the solution of the following system of linear equations:

\begin{equation}
   \begin{cases} A\varphi^0 + B\varphi'^0 = \beta_0 \\ A\varphi^1 + B\varphi'^1 - \log(\bar\alpha) =   A\varphi^{n+1} + B\varphi'^{n+1}\end{cases}
\end{equation}

\noindent The unique solution $(A,B)$ of these equations  is:
\begin{equation}
\begin{split}
        A &= \beta_0 - \frac{\log( \bar \alpha) -\beta_0(\varphi -\varphi^{n+1}) }{\varphi' - \varphi -( \varphi'^{n+1} - \varphi^{n+1})}\\
    B &= \frac{\log( \bar \alpha) -\beta_0(\varphi -\varphi^{n+1}) }{\varphi' - \varphi -( \varphi'^{n+1} - \varphi^{n+1})}
    \end{split}
\end{equation}
    
Thus a closed form solution is obtained that allows to compute the Fibonacci noise schedule ($\beta, \alpha$), given the noise level $\bar \alpha$ and $\beta_0$. 

\subsubsection{Conditioning on Discrete Indexes}
Most DDPMs are not conditioned on the noise level $\bar \alpha$ but instead use a discrete integer index that represents steps in the reverse process.  The two types of conditioning are related.  Instead of giving the network with the exact value $\sqrt{\bar \alpha}$, the network is given the integer $t$ of the interval $[l_t, l_{t-1}]$ (defined in Eq.~\ref{eq:l}), which contains the estimated value.

To enable our method to work with this type of DDPM, we estimate the noise level and feed the network with the integer encoding corresponding interval. 

\subsection{Inference procedure}
Our inference procedure is introduce in algorithm~\ref{alg:alpha_inf}. The idea is to have a set of step indexes $U$ for which, we readjust the noise schedule. For simplicity, in all of our experiments, $U=\{1,2,\dots,N\}$ for a given number of steps $N$.

The adjustment is done using the neural network $P_\theta$, which estimates the noise level $\bar \alpha$. Given this estimation,  we deduce the sets of parameters $\mathbf{\alpha}, \mathbf{\beta}$ for the remaining denoising steps, as shown in the Noise Schedule adjustments Section.   

The noise schedule (vectors $\alpha$ and $\beta$) is initialised with a set of predefined values (function initialNoiseSchedule()). The rest of the algorithm (lines 3-15) is very similar to the algorithm \ref{alg:DDPM_inf}. The only difference (lines 7-9)  is that we use a set of iteration indexes $U$ for which we will adjust the noise schedule vectors.

For the iteration $n\in U$, we estimate the noise level using the model $P_\theta$, then we follow the deterministic method described previously (function updateNoiseSchedule($\bar \alpha, n$)) to compute the adjustment of the noise schedule.

\section{Experiments}
To demonstrate the wide applicability of our approach, we perform experiments in both speech synthesis and image generation. In both cases, we determine the optimal few-step scheduling for state of the art models.

\begin{table}[t]
\centering
\begin{tabular}{lccc}
\toprule
                  & MCD ($\downarrow$)  & PESQ ($\uparrow$) & STOI ($\uparrow$) \\ \midrule
$1000$ iterations & 2.65           & 3.29          & 0.959                   \\ 
Grid Searched     & \textbf{2.76}  & 2.78          & 0.924                   \\ 
Our method    & 2.96  & \textbf{3.14} & \textbf{0.943} \\ \bottomrule
\end{tabular}
\smallskip
\caption{Comparison between a grid searched noise schedule and an adjusting noise schedule for speech generation.}
\label{tab:resmcd}
\end{table}

\begin{figure*}[t]
\centering
\begin{minipage}[c]{0.32\linewidth}
        \includegraphics[width=.99\textwidth,height=.99\textheight,keepaspectratio]{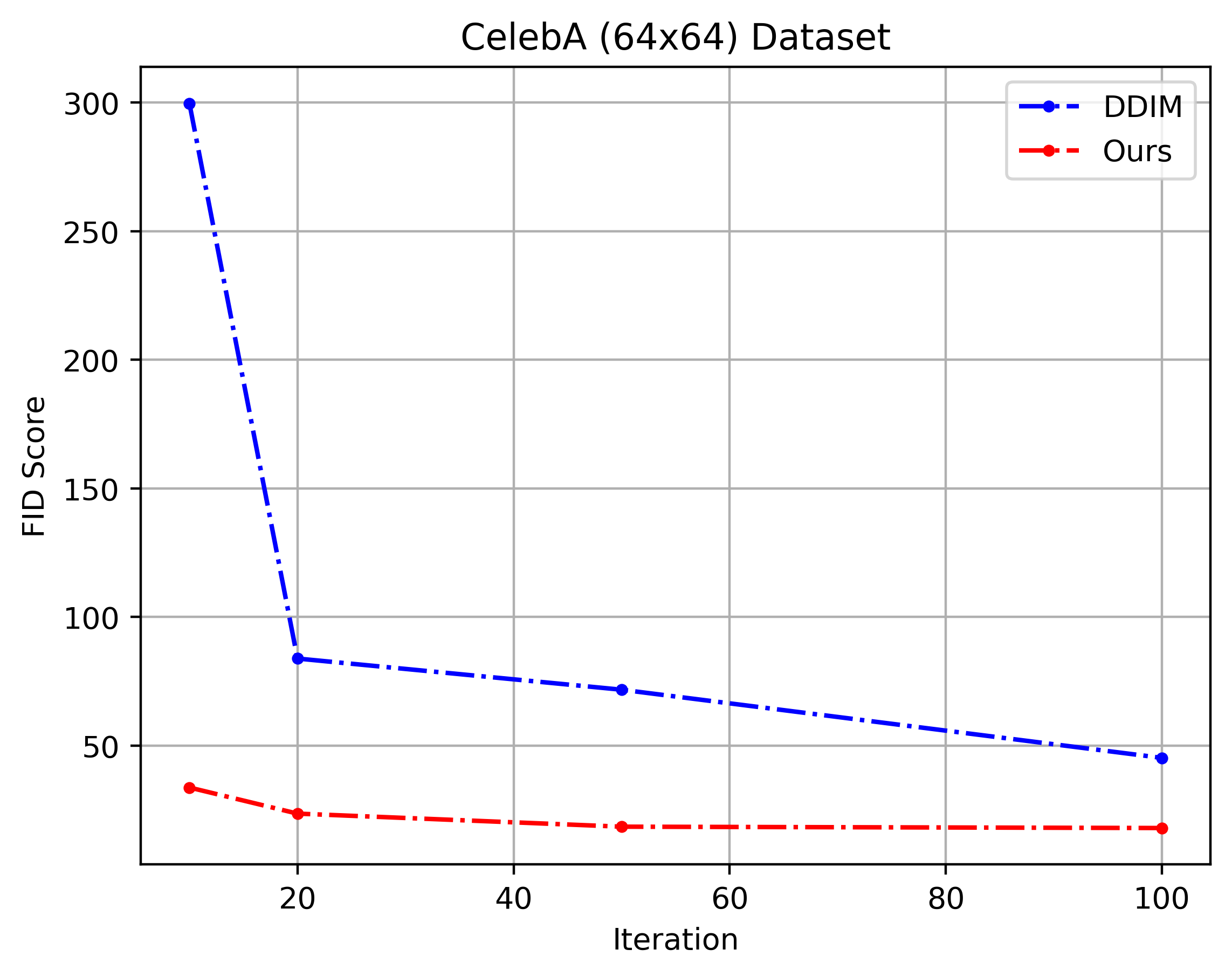}
    \caption{FID Score obtained for CelebA for our method and DDPM as a function of the number of iterations.}
    \label{fig:CelebA_DDPM}
\end{minipage}%
\hfill
\begin{minipage}[c]{0.65\linewidth}
\begin{tabular}{c@{~}c}
\includegraphics[width=.49\textwidth,height=.49\textheight,keepaspectratio]{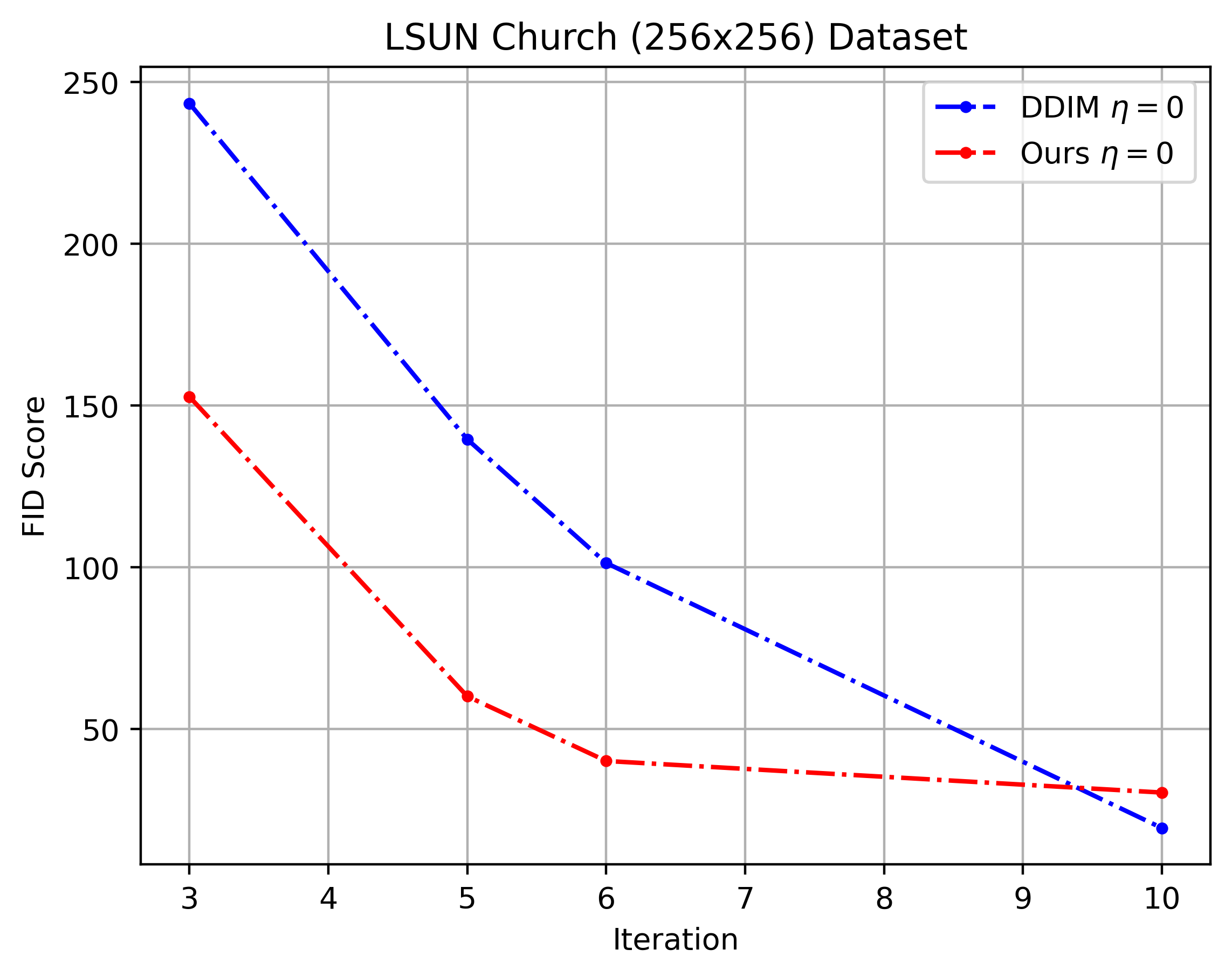}& 
\includegraphics[width=.49\textwidth,height=.49\textheight,keepaspectratio]{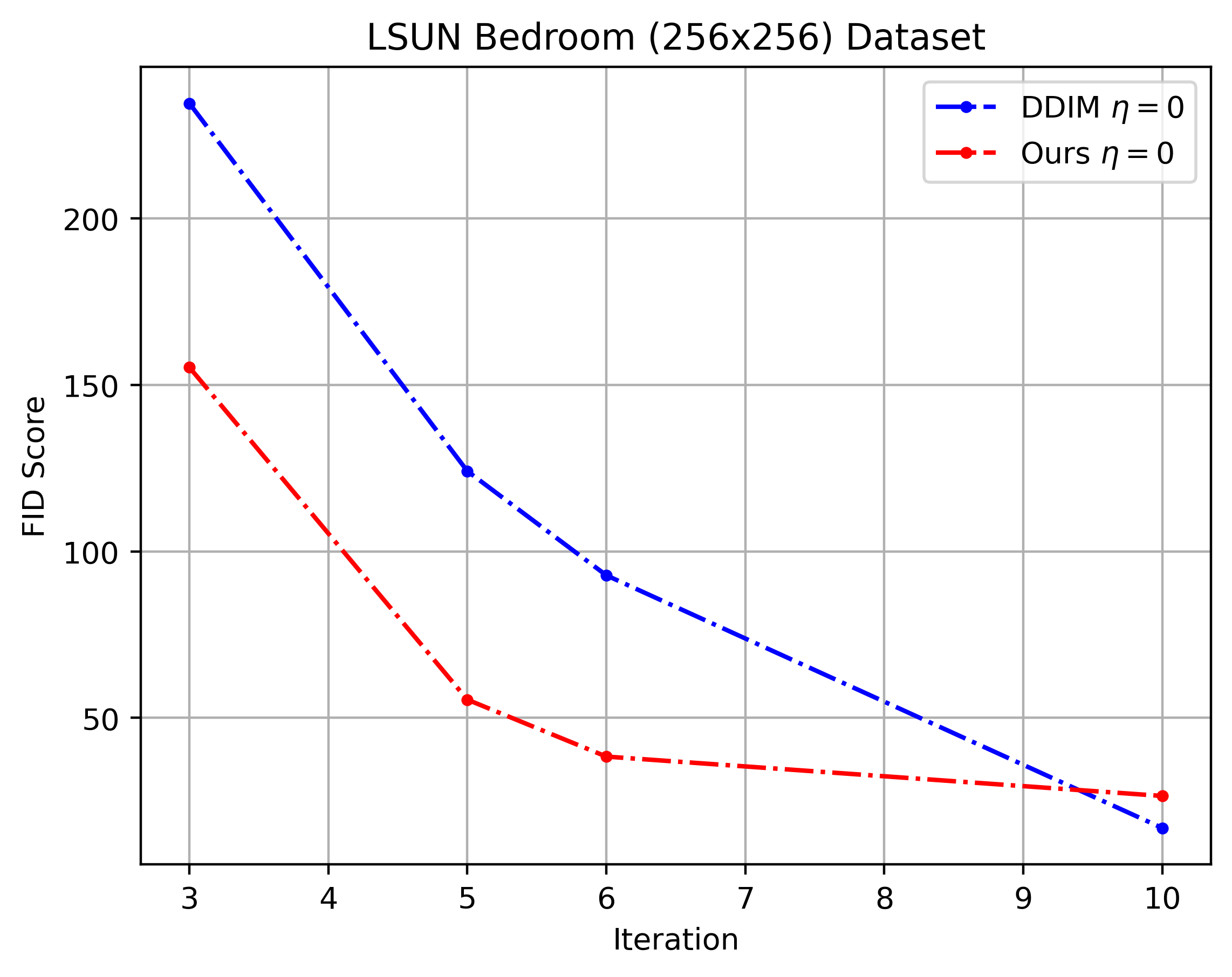} \\
(a) & (b) \\
\end{tabular}
\caption{The FID score, over a small number of iterations, obtained for LSUN (a) Church and (b) Bedroom classes, for DDIM method and for our method.}
\label{fig:FID_DDIM_BEDROOM_CHURCH}
\end{minipage}
\end{figure*}

\subsection{Speech Generation}

For speech generation, we used a WaveGrad implementation that came with a grid searched noise schedule for $6$ iterations \cite{ivangit}. We trained a small model $P_\theta$ based on a the version of ConvTASNet \cite{luo_conv-tasnet_2019} architecture. 

The model is composed of the encoding and masking module of ConvTASNet. Its decoding part is not utilized. The ConvTASNet architecture cuts, using a separator network, the signal in chunks and processes each chunk individually. Our network further applies a fully connected layer with an output dimension of $1$, followed by a sigmoid activation function to each chunk. The final output $\hat \alpha$ is the average of the outputs of all the chunks. The encoder has $N=64$ filters  of size $L=16$ (the parameter names follow~\cite{luo_conv-tasnet_2019}). The separator uses stacks of $X=4$ convolutional blocks of kernel size $P=4$ with $R=4$ repeats. It has $H=128$ channels in the convolutional blocks and $B=64$ in the bottleneck and residual paths. 

To train our model $P_\theta$, we use the same split of the LJ-Speech dataset \cite{ljspeech17} used by \cite{ivangit} for training the WaveGrad model. Our model was trained with the Adam optimizer \cite{Adam} and a learning rate of $0.001$. Each batch includes $32$ samples from an audio duration of $3$ seconds with a sampling rate of $22050$ Hz.

Next, we evaluate the synthesis quality. The DDPM employed in our experiments was trained using Mel Spectrogram as the conditioning signal $x$. Following~\cite{chen_wavegrad_2020}, in our experiments, the ground truth Mel-Spectrogram is given as an input in order to allow the computation of the relative speech metrics (MCD\cite{mcd_paper}, PESQ\cite{PESQ_paper}, STOI\cite{STOI_paper}).  

In our experiments, we performed inference for six iterations with an adjustment of the noise schedule at every step i.e. $U = \{1,\dots 6\}$. This noise schedule adjustment uses the Fibonacci method.

Sample results can be found under the following \href{https://enk100.github.io/Noise-Estimation-for-Generative-Diffusion-Models/}{link}. As can be heard from the samples that are provided, for few iterations, our method obtains a much better improvement than the baseline method, after applying a costly grid search for the optimal parameters of the baseline method. This is also captured by the qualitative results, presented in Tab.~\ref{tab:resmcd}. Even though our method results in a small decrease in the MCD, we demonstrate a large improvement in both PESQ and STOI.

\begin{figure}
    \centering
        \includegraphics[width=.859065\linewidth]{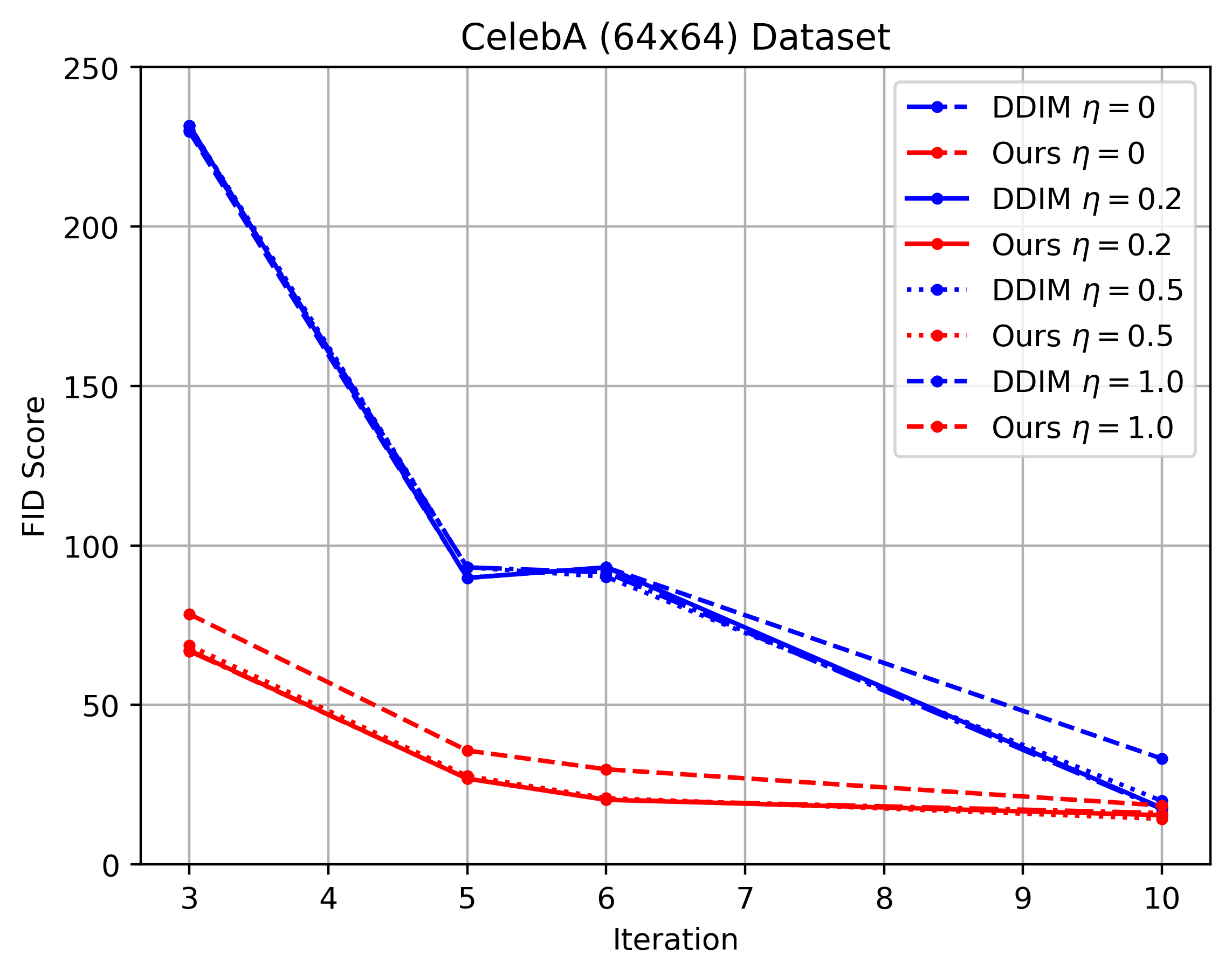}
        \vspace{-3mm}
    \caption{The FID score, over a small number of iterations, obtained for the CelebA dataset, for different $\eta$ values, for DDIM method and for our method.}
    \label{fig:FID_DDIM}
\end{figure}

\subsection{Image Generation}
For image generation, we use denoising diffusion models trained in \cite{DDPMgithub}, relying on the implementation available in \cite{DDIMgithub}. We trained our model $P_\theta$ on three image datasets (i) CelebA 64x64 \cite{liu2015faceattributes}, (ii) LSUN Bedroom 256x256, and (iii) LSUN Church 256x256 \cite{yu15lsun}.

The model $P_\theta$ used for the noise estimation employed a  VGG11~\cite{simonyan2014very} backbone pretrained on ImageNet \cite{deng2009imagenet}. We added ReLU activations, followed with a fully connected layer with an output dimension $1$ and a final sigmoid activation to the backbone. 
An Adam optimizer\cite{Adam} with a learning rate of $10^{-4}$ was used, with a batch size of size 64.

The Fréchet Inception Distance (FID)~\cite{heusel2017gans} is used as the benchmark metric. For all experiments, similarly to the DDIM paper, we compute the FID score with $50,000$ generated images using the torch-fidelity implementation~\cite{torchfidelity}.

Figure~\ref{fig:FID_DDIM} depicts the FID score of our method using the update in Eq.~\ref{eq:DDPM_update} is comparison to the baseline DDIM method. Both methods use the same DDIM diffusion model provided in \cite{DDIMgithub} for CelebA. Our method uses the adjusted noise schedule at every step and generates a linear noise schedule. 

Different plots are given for different values of  $\eta$, which is the parameter that control $\tilde \sigma$ in Eq.~\ref{eq:DDIM_update}:
\begin{equation}
    \tilde \sigma= \eta \sqrt{\beta_n(1 - \bar \alpha_{n-1})(1 - \bar \alpha_n)}
\end{equation}

As can be seen, our method improves by a large gap the FID score the DDIM method. For example, for three iteration with $\eta=0.0$ our method improve the FID score by $163.4$. The gap in performance is maintained for up to 10 iterations. 

\begin{figure}[t]
\centering
\begin{tabular}{c}
\includegraphics[width=.975\linewidth]{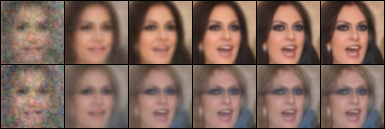}\\ 
\includegraphics[width=.975\linewidth]{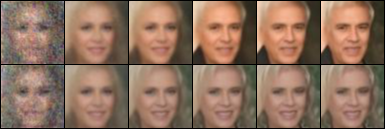}\\
\includegraphics[width=.975\linewidth]{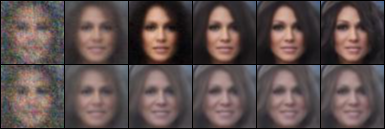} \\
\end{tabular}
\smallskip
\caption{Typical examples of the denoising processes for 6 iterations $\eta=0$. For three different noise inputs, we compare our method (top) to DDIM (bottom).}
\label{fig:evolution_celebA}
\end{figure}

In Figure \ref{fig:evolution_celebA}, we demonstrate the progression over a six iteration denoising process for both our method and the DDIM method that we use as a baseline. {Evidently, our generated images are more coherent and sharper than the DDIM method.}

\begin{figure*}
\centering
\begin{tabular}{c@{~}c@{~}c}
\includegraphics[width=0.32\linewidth]{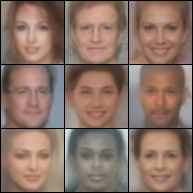}& 
\includegraphics[width=0.32\linewidth]{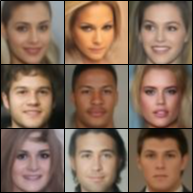}&
\includegraphics[width=0.32\linewidth]{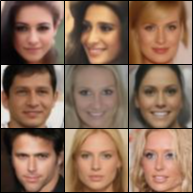}\\
(a) & (b) & (c) \\
\end{tabular}
\smallskip
\caption{Generated images with our method $\eta=0$ for (a) 3, (b) 5 and (c) 6 iterations.}
\label{fig:celebA_generations}
\end{figure*}

\begin{figure*}
\centering
\begin{tabular}{cc}
    \includegraphics[width=0.45\linewidth]{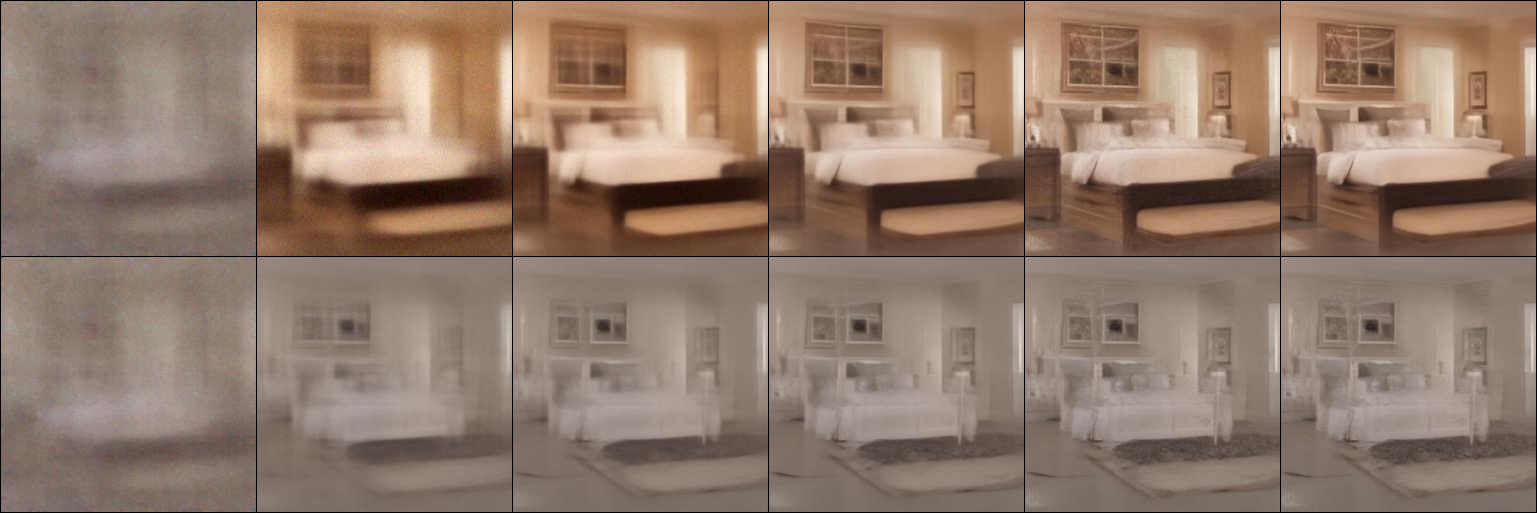} &
    \includegraphics[width=0.45\linewidth]{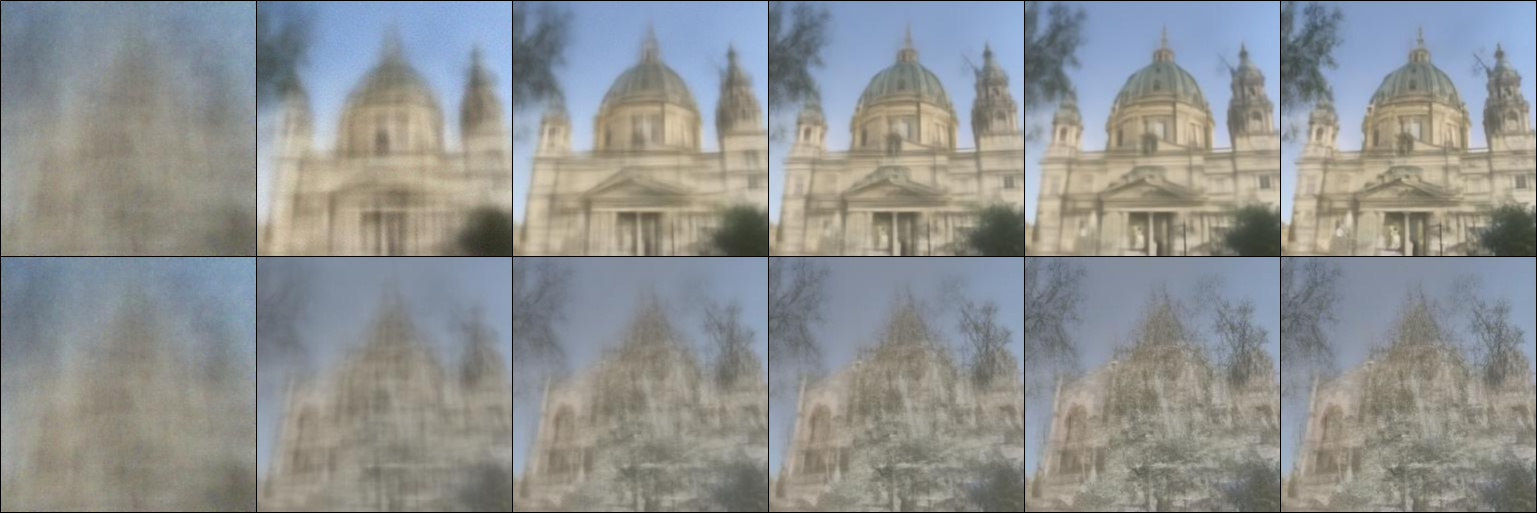} \\
    \includegraphics[width=0.45\linewidth]{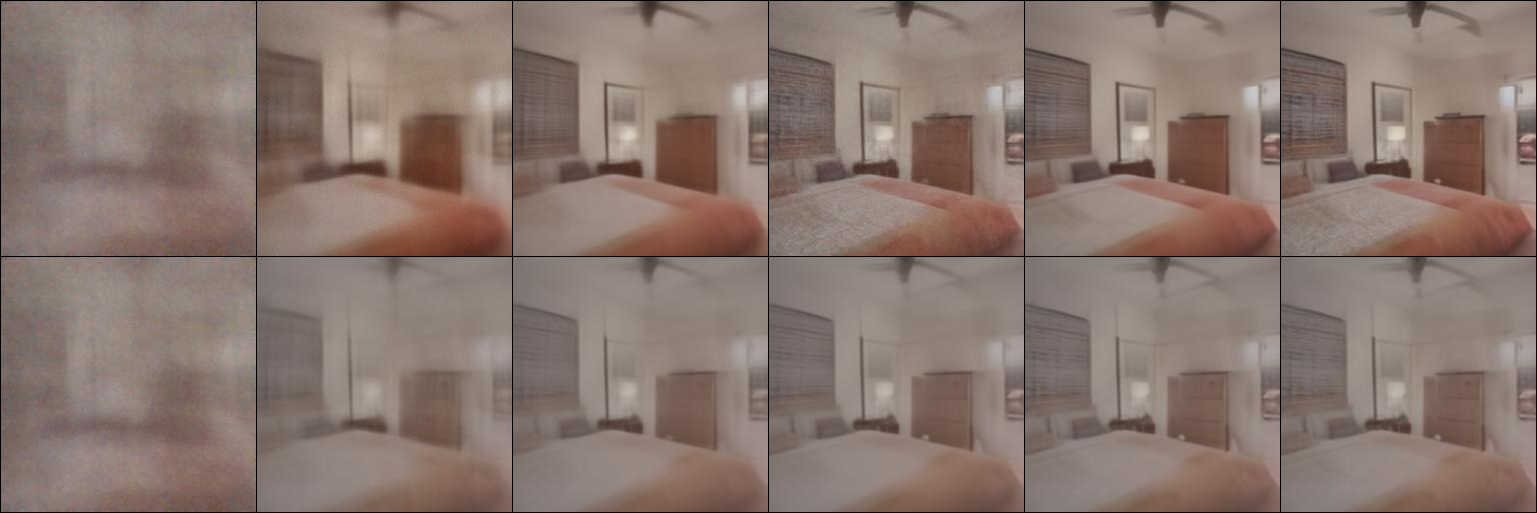} &
    \includegraphics[width=0.45\linewidth]{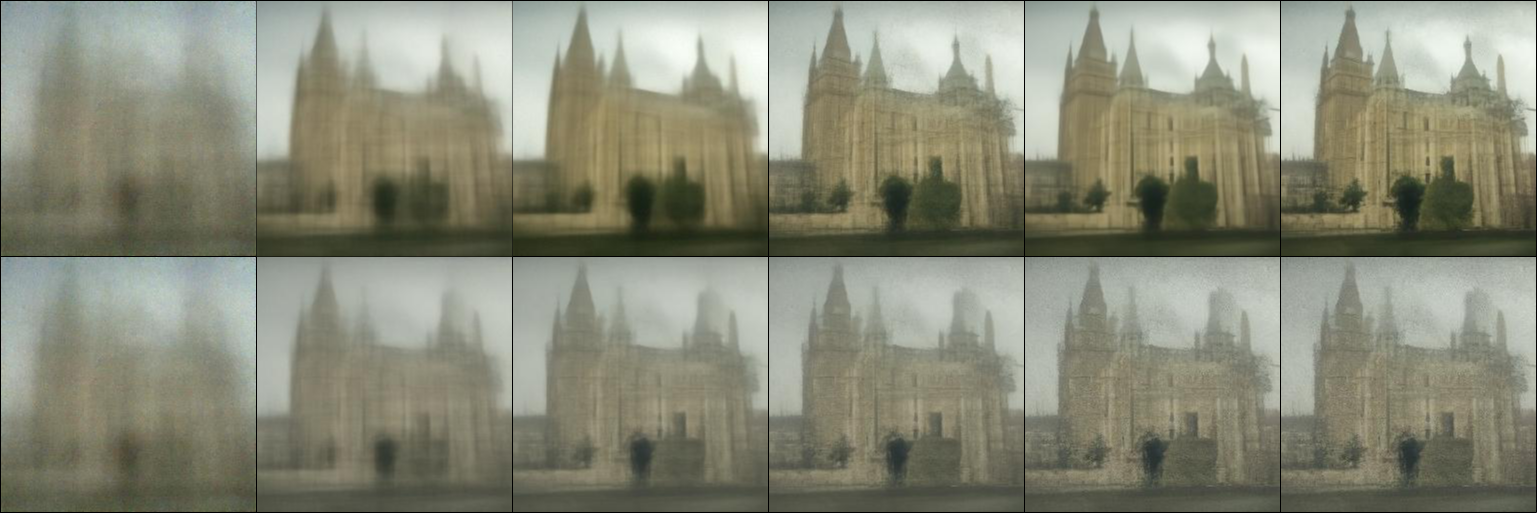} \\
(a) & (b) \\
\end{tabular}
\caption{LSUN 256x256 synthesis examples for $N=6$ iterations. The same input noise used for both process. (a) Church dataset, (b) Bedroom dataset. The top row in each example is our method and the bottom row is the DDIM method. }
\label{fig:ev_bed}
\end{figure*}

\begin{figure*}[t]
\centering
\begin{tabular}{c@{~}c@{~}c@{~}c}
\includegraphics[width=0.2325\linewidth]{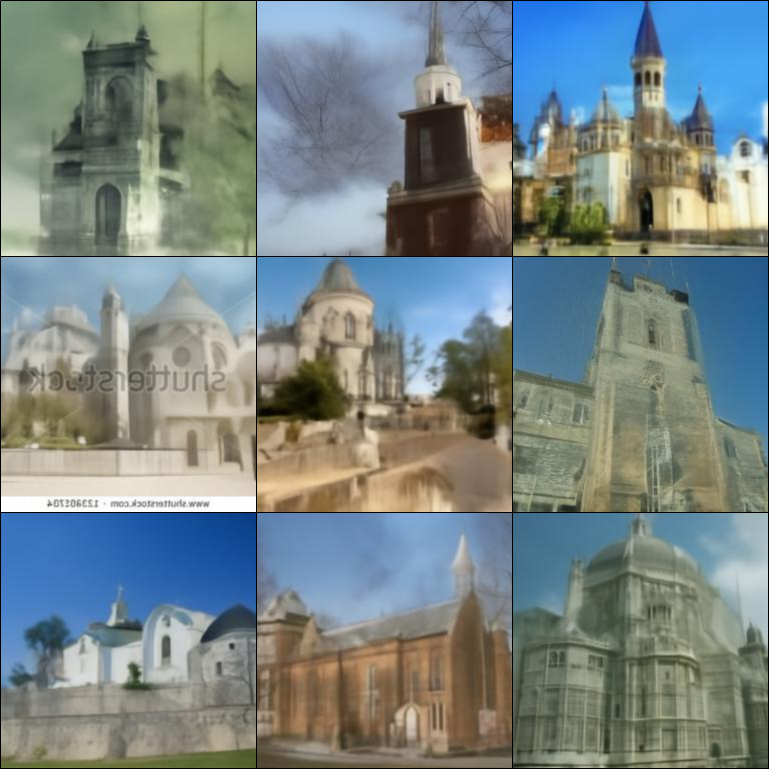}& 
\includegraphics[width=0.2325\linewidth]{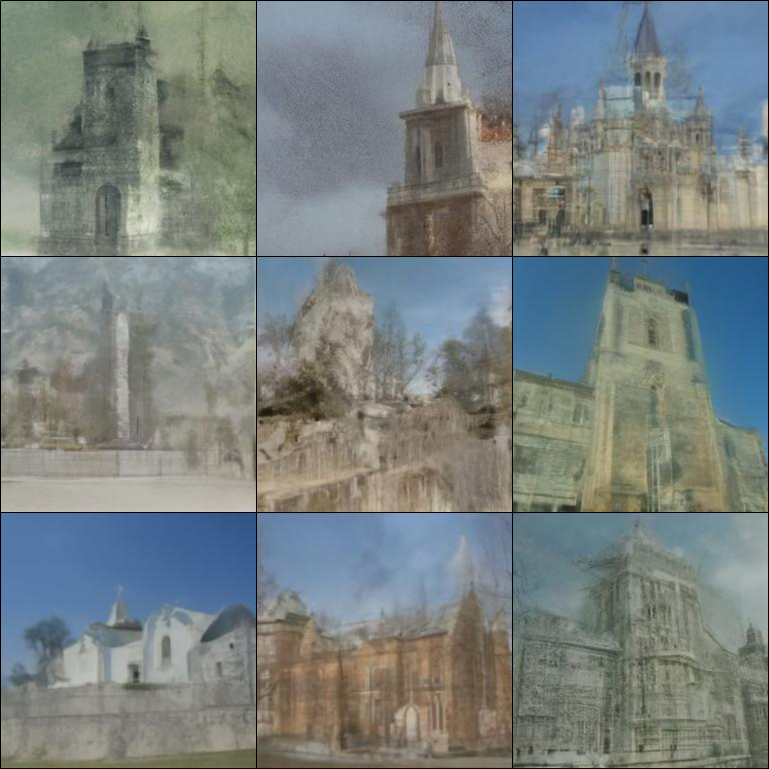}& 
\includegraphics[width=0.2325\linewidth]{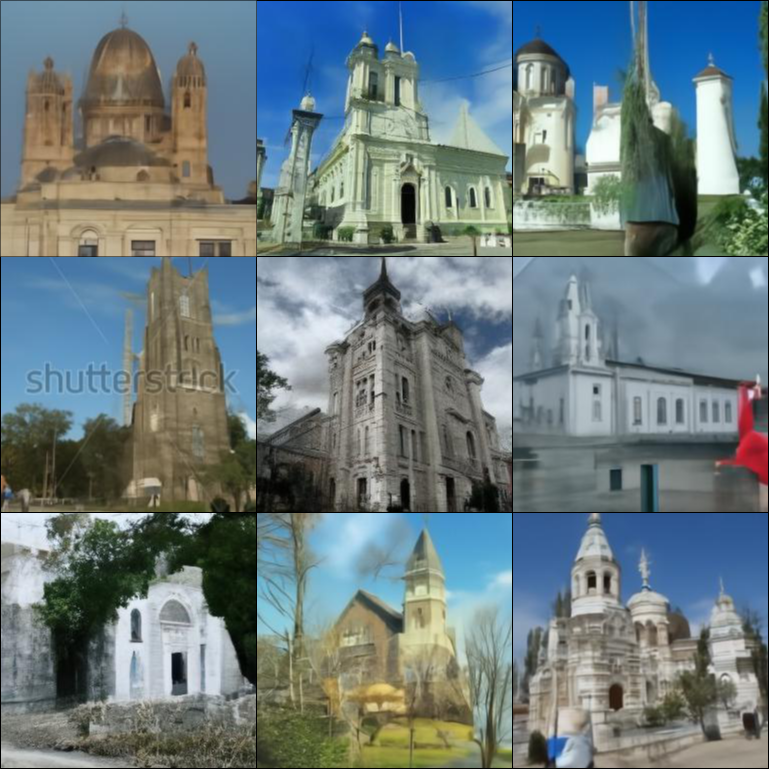}& 
\includegraphics[width=0.2325\linewidth]{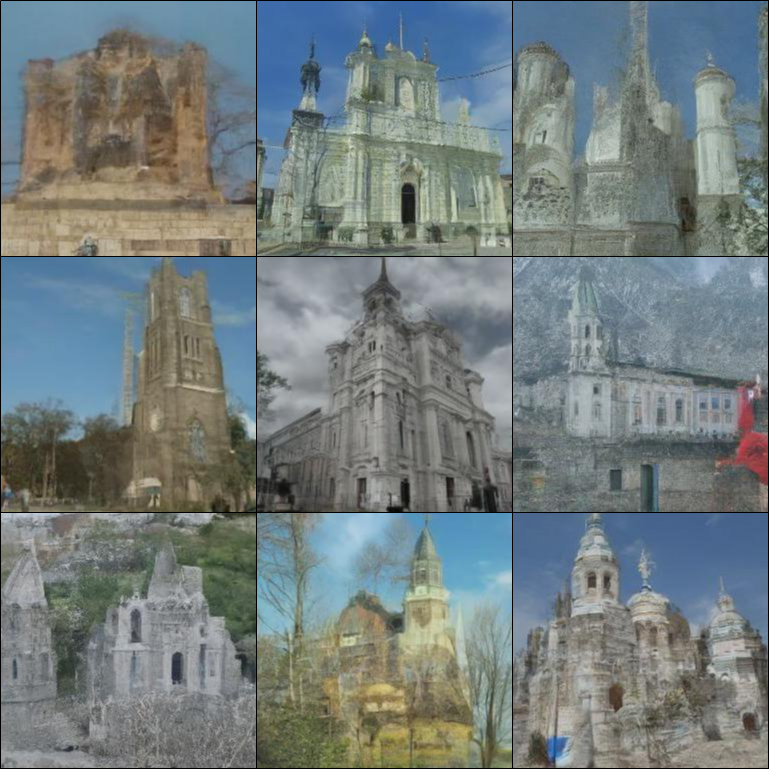}\\ 
(a) & (b) & (c) & (d)\\ \\

\includegraphics[width=0.2325\linewidth]{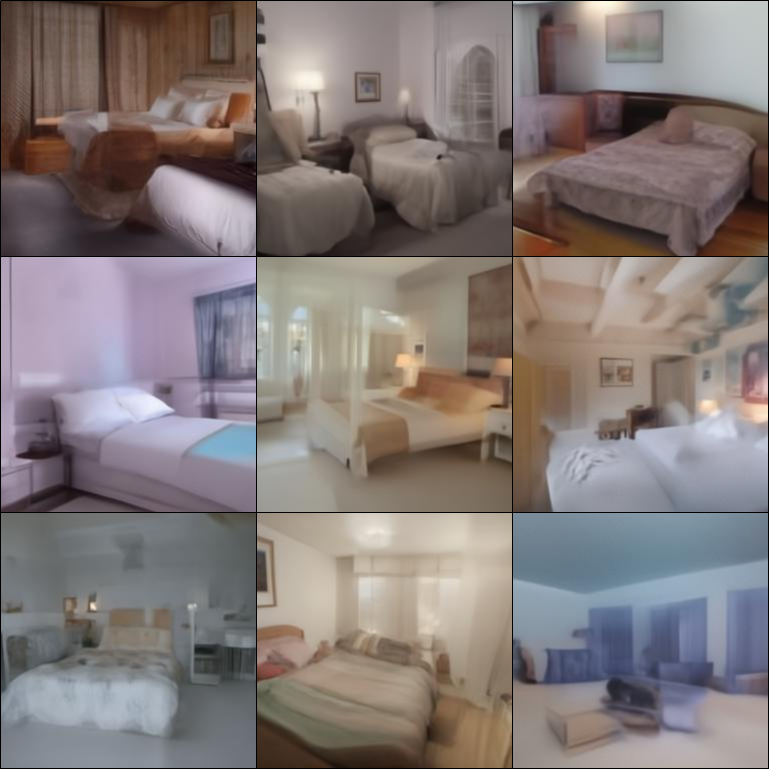}& 
\includegraphics[width=0.2325\linewidth]{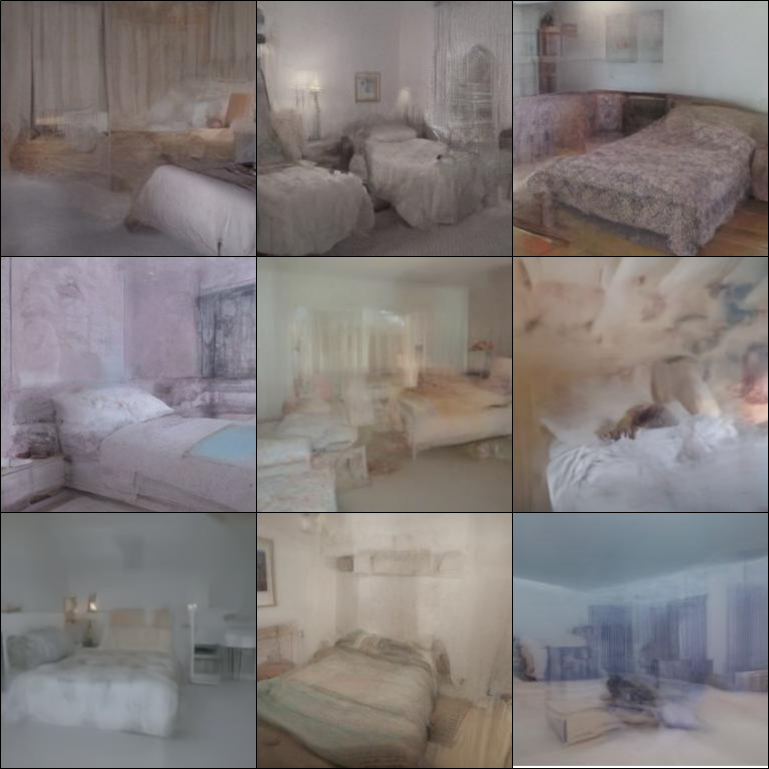}& 
\includegraphics[width=0.2325\linewidth]{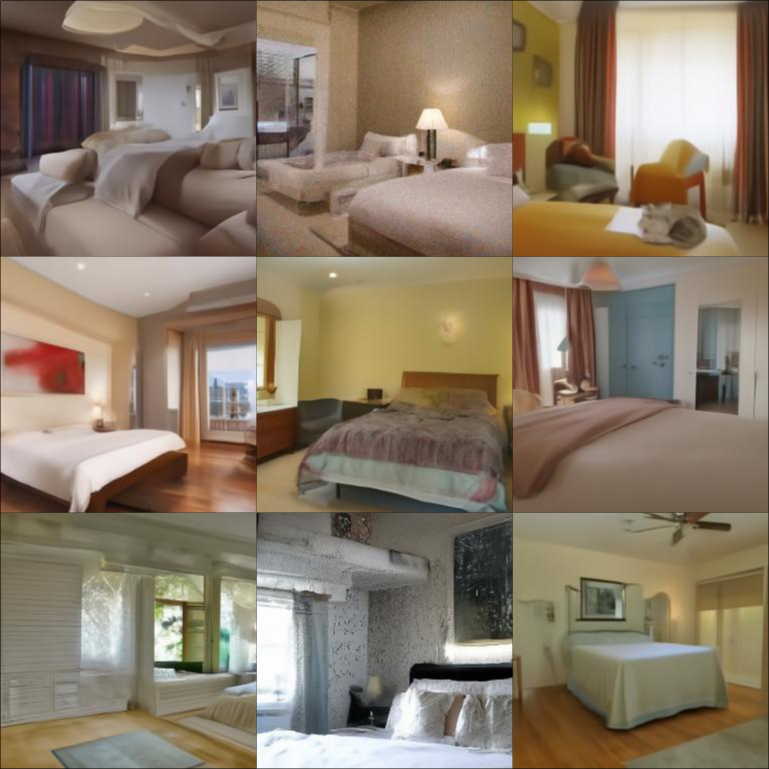}& 
\includegraphics[width=0.2325\linewidth]{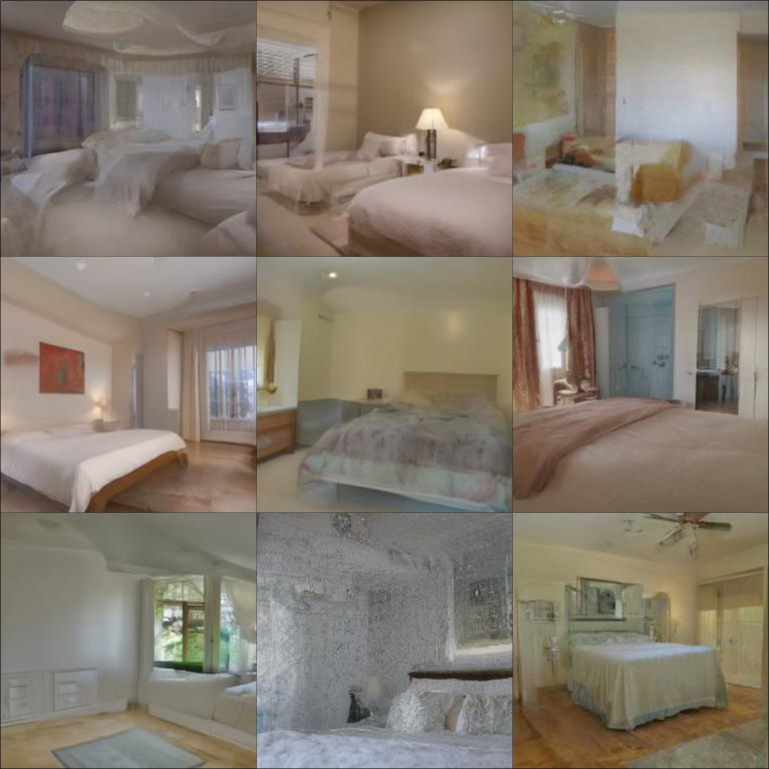}\\
(e) & (f) & (g) & (h)\\
\end{tabular}
\caption{LSUN Church and Bedroom 256x256 datasets. Comparison of our method and the DDIM baseline for $6$ and $10$ iterations. First row is for LSUN Church 256x256 (a) ours with $6$ iterations, (b) DDIM  with $6$ iteration, (c) ours with $10$ iterations, (d) DDIM with $10$ iterations. Second row is for LSUN Bedroom 256x256 (e) ours with $6$ iterations, (f) DDIM with $6$ iterations, (g) ours with $10$ iterations, (h) DDIM with $10$ iterations.}
\label{fig:comparison}
\end{figure*}

Figure \ref{fig:celebA_generations} presents samples for celebA 64x64 generation, given a different number of target steps. Our inference procedure is able to generate decent images for as little as $3$ denoising steps. We also show convincing generations results for $5$ and $6$ steps generation processes. This results demonstrate that our method helps toward finding the best non-Markovian denoising process, especially when generating with very few iterations. Figures \ref{fig:ev_bed} and \ref{fig:comparison} depicts comparison of generative processes and generated samples between our method and the DDIM baseline. Our method overall clearly improves sharpness and contrast over the baseline. 


\begin{figure*}
\centering
\begin{tabular}{cc}
    \includegraphics[width=.45\linewidth]{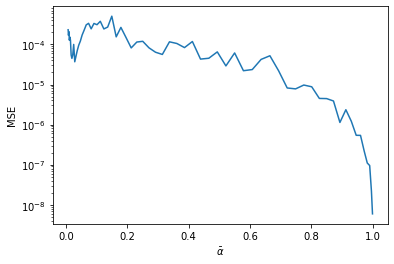} &
    \includegraphics[width=.5\linewidth]{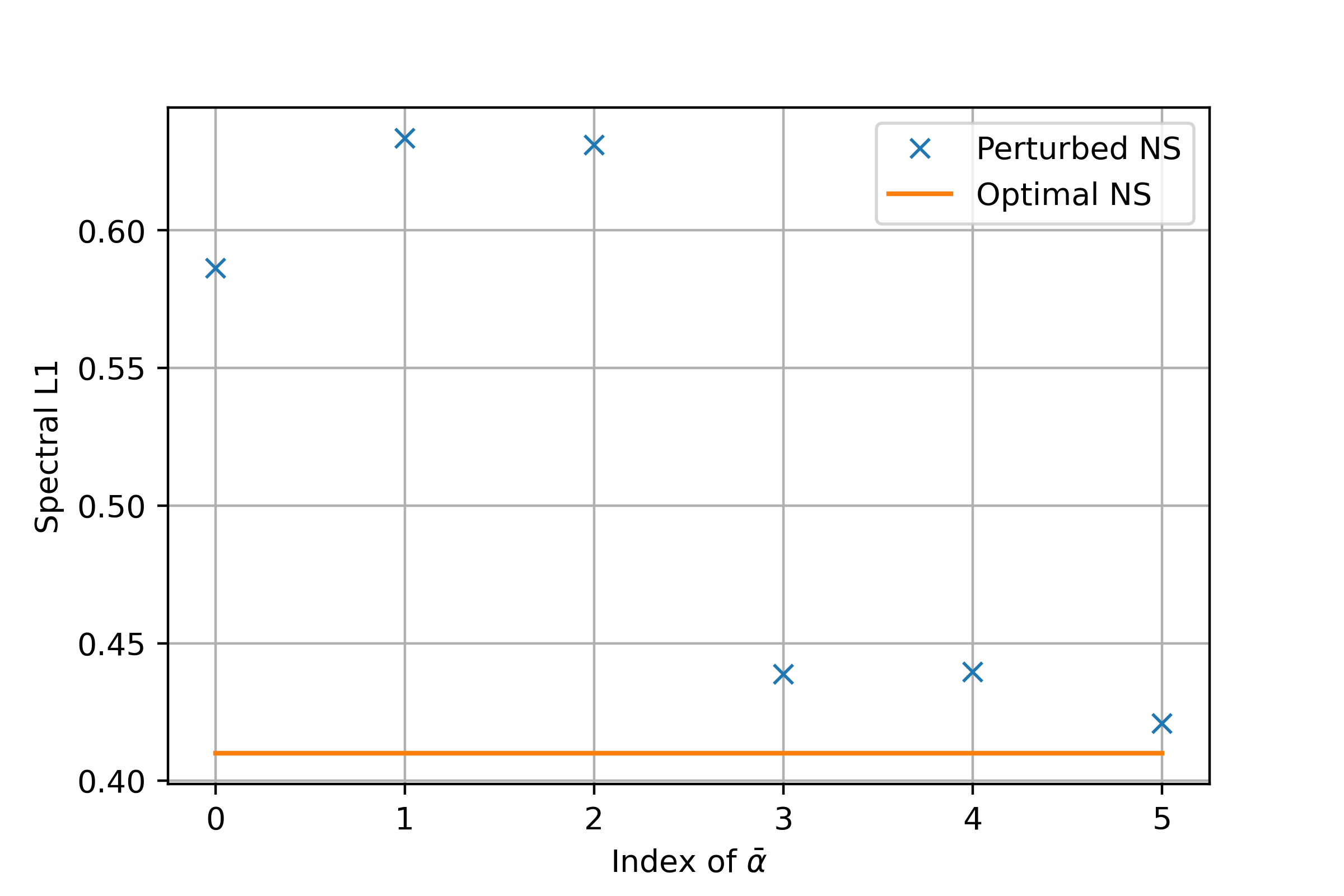} \\
(a) & (b) \\
\end{tabular}
\caption{(a) Performance of the network $P_\theta$ for the speech data for the noise estimation task itself, (b) L1 distance between the mel spectrogram of the groundtruth and the generated sample w.r.t the index of the perturbed}
\label{fig:p_theta_and_l1}
\end{figure*}

\begin{table}[t]
\centering
\begin{tabular}{l@{~~}c@{~~}c@{~~}c@{~~}c@{~~}c@{~~}c}
\toprule
$|U|$(\# adjustments) & 5    & 3    & 2    & 1    & 0    & Grid Search \\
\midrule 
PESQ                  & 3.14 & 3.11 & 3.02 & 2.61 & 2.54 & 2.78 \\
\bottomrule
\end{tabular}
\caption{PESQ score for six-iteration generations with respect to the number of noise schedule adjustments.}
\label{tab:adjustments}
\end{table}

\begin{table}[t]
\centering
\begin{tabular}{@{}l@{~}ccccc@{}}
\toprule
Method/\#Iteration  & 10 & 20 & 50 & 100        \\ \midrule
No adjustment (DDPM/DDIM)          & 0.41  & 0.80  & 2.02  & 4.03          \\ 
Adjustment every iteration    & 0.59  & 1.19  & 2.92  & \textit{6.02} \\ 
\bottomrule
\end{tabular}
\caption{Mean image generation time in second obtained for CelebA 64x64 for our method and DDPM/DDIM.}
\label{tab:Time_celebA}
\end{table}

In Figure \ref{fig:CelebA_DDPM}, we compare our method with DDPM for the CelebA dataset. As can be seen, our method improves, by a large margin, the results of the DDPM method up to 100 iterations. For example, for $10$ iteration, we improve the FID score by $266.03$. 

Similarly, in Figures \ref{fig:FID_DDIM_BEDROOM_CHURCH} we provide the FID score for LSUN Church and Bedroom 256x256 datasets, using the published DDIM models. As can be seen for a small number of iteration, i.e. less then $10$, our method greatly improves the results of the baseline method DDIM. In \cite{DBLP:journals/corr/abs-2101-02388}, authors propose a distilation framework that can sample in one iteration. The method performs FID scores of $54.09$ on LSUN Church and $60.97$ on Bedroom. As can be seen our method gets better results for as few as 6 iterations. 

\paragraph{Additional Experiments}\label{sec:AddExp}
We perform ablation experiments to justify some claims and choice in our method. 
In Figure ~\ref{fig:p_theta_and_l1}(b) we computed the optimal noise schedule for a data sample. We perturbed the values of $\bar \alpha_t$ for the different $t$ by $10^{-4}$. One can clearly see that perturbations on $\bar \alpha_0$, $\bar \alpha_1$, $\bar \alpha_2$ results in huge drop of the performance whereas the others do not change the results much. It shows that the performance of the sampling algorithm is mostly dependent on the last steps, where the alpha values are small. This is why we use the loss \ref{eq:loss} and want our model to perform his best when $\bar \alpha$ is close to $1$.
In order evaluate the accuracy of $P_\theta$ in recovering $\bar \alpha$, noisy speech signals are generated according to Eq.~\ref{eq:closedy_n} with $\bar \alpha$ values between 0 and 1. The noise level is then estimated by the trained network $P_\theta$. 
The results of this experiment are presented in Figure~\ref{fig:p_theta_and_l1}(a). The Mean Square Error (MSE) between the ground truth $\bar \alpha$ (used to generate the noisy speech) and the network estimation $\hat{\alpha}$ is depicted for $\bar \alpha$ between 0 and 1. Each point is computed with 16 audio files of 3 secs from the validation set. As can be seen, our model is able to estimate the noise level within an error of at most $10^{-4}$ and with even better performance when alpha is close to $1$, which is where the precision is critical to the performance of our method.
In Table~\ref{tab:adjustments} we provide the PESQ score for $6$ denoising steps, with various number of adjustments. As can been seen, the best results are when readjusting every steps, yet our method already outperforms the grid search for two adjustments.
Table~\ref{tab:Time_celebA} depicts the mean run-time to generate an image on an Nvidia Titan X over 1000 generations. It compares our method that adjusts the noise schedule at every steps with a fully predefined schedule. This table shows that even though we use VGG11 which is not a cost efficient method to estimate the noise, the generation time increases by a moderate amount.

\section{Conclusions}
When employing diffusion models with a limited number of steps, it is required to carefully choose the schedule of the synthesis process. Some of the previous methods perform a grid search to find the optimal global noise parameters per each number of steps. However, this fixed selection does not adjust according to the specific sample that is being generated. Our method adjusts on-the-fly the noise parameters and thus alters the subsequent states of the process. Our solution is based on estimating, during inference, the current noise level. It is, therefore, generic and independent, given the current sample, from the conditioning parameters. It remains for future work to check whether the same $P_{\theta}$ network can be used across multiple datasets.
\clearpage
\bibliography{aaai22}
\end{document}